\documentclass[letterpaper, 10 pt, conference]{ieeeconf}  
\usepackage[utf8]{inputenc}
\usepackage{graphicx} 
\usepackage{svg}
\usepackage{cite}
\usepackage{amsmath,amssymb,amsfonts}
\usepackage{algorithmic}
\usepackage{textcomp}
\usepackage{xcolor}
\usepackage{booktabs}
\usepackage{algorithm}
\usepackage{url}
\usepackage{multirow}
\usepackage{svg}
\usepackage[acronym, nonumberlist, nogroupskip]{glossaries}
\makeglossaries

\newacronym{mse}{MSE}{Mean Squared Error}
\newacronym{mae}{MAE}{Mean Absolute Error}
\newacronym{rmse}{RMSE}{Root Mean Squared Error}
\newacronym{psg}{PSG}{polysomnography}
\newacronym{shap}{SHAP}{SHapley Additive exPlanations}
\newacronym{lstm}{LSTM}{Long Short-Term Memory}
\newacronym{cnn}{CNN}{Convolutional Neural Network}
\newacronym{rnn}{RNN}{Recurrent neural network}
\newacronym{loocv}{LOOCV}{Leave-one-out Cross-Validation}
\newacronym{tta}{TTA}{Test-Time Adaptation}
\newacronym{bmi}{BMI}{Body Mass Index}

\IEEEoverridecommandlockouts                              

\title{\LARGE \bf
Individualized and Interpretable Sleep Forecasting via a two-stage Adaptive Spatial-Temporal Model}

\author{Xueyi Wang$^{1}$, Claudine J. C. Lamoth$^{2}$, and Elisabeth Wilhelm$^{3}$
\thanks{*This work was supported by the Healthy Living as a Service Project}
\thanks{$^{1}$Xueyi Wang is with the Engineering and Technology Institute Groningen, University of Groningen, Groningen, Netherlands.
        {\tt\small xueyi.wang@rug.nl}}%
\thanks{$^{2}$Claudine J. C.  Lamoth with the Department of Human Movement Sciences, University Medical Center Groningen,
        Groningen, Netherlands.
        {\tt\small c.j.c.lamoth@umcg.nl}}%
\thanks{$^{3}$Elisabeth Wilhelm with the Engineering and Technology Institute Groningen, University of Groningen,
        Groningen, Netherlands.
        {\tt\small e.wilhelm@rug.nl}}%
}

\begin{document}

\maketitle
\thispagestyle{empty}
\pagestyle{empty}

\begin{abstract}

Sleep quality impacts well-being. Therefore, healthcare providers and individuals need accessible and reliable forecasting tools for preventive interventions. This paper introduces an interpretable, individualized adaptive spatial-temporal model for predicting sleep quality. We designed a hierarchical architecture, consisting of parallel 1D convolutions with varying kernel sizes and dilated convolution, which extracts multi-resolution temporal patterns—short kernels capture rapid physiological changes, while larger kernels and dilation model slower trends. The extracted features are then refined through channel attention, which learns to emphasize the most predictive variables for each individual, followed by bidirectional \gls{lstm} and self-attention that jointly model both local sequential dynamics and global temporal dependencies. Finally, a two-stage adaptation strategy ensures the learned representations transfer effectively to new users. We conducted various experiments with five input window sizes (3, 5, 7, 9, and 11 days) and five prediction window sizes (1, 3, 5, 7, and 9 days). Our model consistently outperformed time series forecasting baseline approaches, including \gls{lstm}, Informer, PatchTST, and TimesNet. The best performance was achieved with a three-day input window and a one-day prediction window, yielding a \gls{rmse} of 0.216. Furthermore, the model demonstrated good predictive performance even for longer forecasting horizons (e.g, with a 0.257 \gls{rmse} for a three-day prediction window), highlighting its practical utility for real-world applications. We also conducted an explainability analysis to examine how different features influence sleep quality. These findings proved that the proposed framework offers a robust, adaptive, and explainable solution for personalized sleep forecasting using sparse data from commercial wearable devices. 
\end{abstract}


\section{Introduction}

Sleep plays a critical role in physical and mental health, and poor sleep quality is associated with a range of health risks, including cardiovascular disease, cognitive decline, and depression~\cite{grandner2022sleep}. Most existing research on sleep assessment has focused on classification of sleep stages or retrospective sleep quality estimation, often relying on controlled laboratory data such as \gls{psg}~\cite{rosenberg2013aasm}. Studies with \gls{psg} provide high-fidelity insights, but  \gls{psg} are costly, time-consuming, and impractical to scale for large populations. Furthermore, most studies do not address the predictive task of forecasting sleep quality, which is essential for proactive health interventions. Moreover, in data sets that do only rely on a single night of recording, the first-night effect might alter symptoms. 

Wearable sensors provide an alternative to traditional polysomnography for sleep monitoring because they allow sleep forecasting applications to expand beyond clinical (laboratory) environments, enabling personalized health monitoring. Wearable sensors such as ECG, accelerometers~\cite{wu2025human}, Radar~\cite{wu2025non}, photoplethysmography, and heart rate monitors embedded in commercial devices like smartwatches, Garmin, and Fitbit provide promising alternatives for unobtrusive long-term sleep monitoring. Commercial smartwatches demonstrate reliability for basic sleep/wake detection with an average accuracy of 87.2\% across 53 devices, consistently performing in the 85-96\% range across multiple studies and maintaining this accuracy in both laboratory and real-world home environments~\cite{birrer2024evaluating}. Fitbit devices demonstrated 87.81\% sensitivity in detecting sleep periods and achieved 92.97\% sensitivity with 52.44\% specificity when operating in 4-stage mode, indicating reliable sleep detection capabilities~\cite{moreno2019validation}. These devices have demonstrated reliability that makes them valuable tools for both personal health management and scientific research~\cite{birrer2024evaluating, moreno2019validation}. Using data acquired by acti-watches, \gls{cnn} and \gls{lstm} networks, a time-batched version of TB-LSTM, these systems, combining wearables and machine learning, can analyze large volumes of data to achieve an 0.892 accuracy~\cite{sathyanarayana2016sleep}. Moreover, systems using objective sleep data have outperformed subjective self-reports in forecasting cognitive performance~\cite{Winslow2017}. These innovations in wearable sensors to collect large amounts of data in daily life environments, in combination with data analysis, facilitate individualized feedback, early identification of sleep disorders~\cite{Patanaik2018}, and population-level studies crucial for informing public health strategies. For instance, smartphone-derived data has demonstrated how biological and cultural factors influence sleep patterns~\cite{Walch2016}.

Several validation studies have compared the performance of smartwatches against gold-standard methods like \gls{psg} and actigraphy. Fitbit devices, equipped with advanced sensors including accelerometers, optical heart rate monitors, and SpO$_2$ sensors, can detect sleep and wake periods, with studies reporting agreement rates over 85\% between \gls{psg} and smartwatch for total sleep time and sleep efficiency~\cite{moreno2019validation, tedesco2019validity}. Commercial wearables like Fitbit and Garmin enable large-scale sleep research through affordability, cloud storage, and ease of use. Their non-invasive nature improves compliance over traditional \gls{psg} methods for longitudinal studies~\cite{purta2016experiences}. Deep learning models mentioned above~\cite{sathyanarayana2016sleep} applied a "black-box" nature, which poses significant limitations in clinical utility and user trust. Especially in healthcare applications, interpretability is crucial for validating predictions, guiding treatment decisions, and ensuring fairness and reliability. Explainable artificial intelligence (XAI) methodologies aim to bridge this gap by providing transparent insights into model decisions and elucidating the importance of various input features~\cite{sindiramutty2024explainable}.

Despite this abundance of wearable-generated data, forecasting of sleep quality or sleep scores is not often explored in scientific literature, especially using data from commercial devices. In this work, we propose a novel adaptive, individualized, and explainable framework for forecasting sleep quality scores using multimodal time-series data collected from a commercial smartwatch (Garmin Vivosmart 5). Our model employs a two-stage domain adaptation strategy to generalize across different users and function effectively in real-world environments where labeled data is limited and individual data distributions differ.


\section{Related Work}

Contemporary research leverages multimodal, high-resolution data streams, including physiological signals such as EEG, heart rate variability, respiratory patterns, behavioral, and environmental data~\cite{Sadeh2015}. Despite such advances, individual differences and the complex interaction between sleep-wake cycles and sleep pressure create ongoing challenges~\cite{Borbely2016}.

Machine learning techniques have substantially improved the accuracy and utility of sleep forecasting. Deep learning models, such as \gls{cnn}, \gls{rnn}, and attention-based\cite{Phan2019} architectures, \gls{cnn}-transformer~\cite{zhang2024ctcnet}, excel at capturing temporal dependencies and modeling complex interactions in sleep data~\cite{Zhao2021}. 

Nevertheless, generalization across diverse populations and environments remains challenging. Traditional machine learning models typically assume identical distributions between training and testing data, an assumption often violated in practical scenarios due to inter-subject variability and different data collection conditions. Transfer learning has enhanced the generalization capabilities of these models across diverse populations and heterogeneous data sources~\cite{peng2025heterogeneity}. Moreover, domain adaptation techniques have emerged as a promising solution, facilitating learning of domain-invariant features and reducing distributional discrepancies~\cite{jia2021multi}. Domain adaptation is crucial in sleep forecasting due to significant variations across populations, environments, and wearable technologies that cause models trained on specific groups or devices to generalize poorly. These variations include physiological differences in heart rate variability, circadian rhythms, and sleep architecture, environmental factors such as climate, noise, and lighting that influence sleep quality~\cite{Obradovich2017}, and technological heterogeneity from different manufacturers' measurement algorithms~\cite{Winslow2017}. Domain adaptation methods address these challenges by transferring knowledge between domains (populations or devices), enabling personalized and reliable predictions essential for clinical and individual sleep management. 
 

\begin{table*}[htbp]
    \centering
    \caption{Acronyms, units, ranges, anomaly, and missing data rates for variables from the Garmin Watch}
    \label{tab:acronyms}
    \scalebox{1}{
    \begin{tabular}{l@{\hspace{1.5cm}}l@{\hspace{1.5cm}}l@{\hspace{1.5cm}}l@{\hspace{1.5cm}}l}
        \toprule
        Acronym & Description & Unit & Anomaly & Missing\\
        \midrule
        TK & TotalKilocalories & kcal & 15 (0.98\%) & 1 (0.09\%) \\
        TS & TotalSteps & arb. & 56 (3.65\%) & 1 (0.09\%) \\
        TD & TotalDistanceMeters & meter & 65 (4.23\%) & 1 (0.09\%) \\
        HA & HighlyActiveSeconds & seconds & 34 (2.21\%) & 1 (0.09\%) \\
        AS & ActiveSeconds & seconds & 27 (1.76\%) & 1 (0.09\%) \\
        MI & ModerateIntensityMinutes & minutes & 51 (3.32\%) & 1 (0.09\%) \\
        RH & RestingHeartRate & beats per minute & 59 (3.84\%) & 1 (0.09\%) \\
        MH & minAvgHeartRate & beats per minute & 34 (2.21\%) & 2 (0.17\%) \\
        XH & maxAvgHeartRate & beats per minute & 19 (1.24\%) & 2 (0.17\%) \\
        AWR & AvgWakingRespirationValue & breaths per minute & 104 (6.67\%) & 2 (0.17\%) \\
        HRV & HighestRespirationValue & breaths per minute & 51 (3.32\%) & 2 (0.17\%) \\
        LRV & LowestRespirationValue & breaths per minute & 117 (7.62\%) & 2 (0.17\%) \\
        DS & deepSleepSeconds & seconds & 25 (1.63\%) & 7 (0.61\%) \\
        LS & lightSleepSeconds & seconds & 33 (2.15\%) & 7 (0.61\%) \\
        RS & remSleepSeconds & seconds & 26 (1.69\%) & 40 (3.47\%) \\
        AW & awakeSleepSeconds & seconds & 54 (3.52\%) & 7 (0.61\%) \\
        AC & awakeCount & arb. & 27 (1.76\%) & 7 (0.61\%) \\
        SS & avgSleepStress & arb. & 48 (3.12\%) & 7 (0.61\%) \\
        RM & restlessMomentCount & arb. & 19 (1.24\%) & 2 (0.17\%) \\
        LR & lowestRespiration & breaths per minute & 80 (5.21\%) & 6 (0.52\%) \\
        HR & highestRespiration & breaths per minute & 65 (4.23\%) & 6 (0.52\%) \\
        AR & averageRespiration & breaths per minute & 78 (5.08\%) & 6 (0.52\%) \\
        H & Hydration & milliliter & N/A & N/A\\
        \textbf{SS} & \textbf{sleep\_overallScore} & \textbf{arb.} & \textbf{24 (1.56\%)} & \textbf{7 (0.61\%)} \\
        \bottomrule
    \end{tabular}
    }
\end{table*}

\gls{tta} is an emerging inference‐time strategy that dynamically refines a pre‐trained model using incoming test data to address distributional shifts common in sleep forecasting. This is particularly valuable for handling inter‐ and intra‐individual variations in sleep patterns due to lifestyle changes or stress, as well as device heterogeneity arising from diverse data collection algorithms~\cite{Sun2020}. Moreover, by adapting in real time to environmental factors such as ambient noise or temperature fluctuations, \gls{tta} enhances the robustness and reliability of sleep‐quality predictions in real‐world settings~\cite{Liang2022}.

Recent studies have integrated XAI into sleep forecasting systems. In clinical settings, understanding contributing factors such as heart rate variability conditions enables clinicians to validate predictions against medical knowledge and make informed treatment decisions. For patients, interpretability increases engagement and adherence with behavioral interventions by clearly linking actions (e.g., screen time) to sleep outcomes~\cite{Ribeiro2016, Samek2019}. Additionally, XAI facilitates model improvement by identifying influential features and addressing biases, ensuring equitable and effective predictions across diverse populations~\cite{Lipton2018}. Thus, XAI transforms sleep forecasting models from black-box predictors into transparent, actionable tools for personalized sleep management. Ahadian et al.~\cite{Ahadian2024} developed a trustworthy forecasting system employing Temporal Convolutional Networks (TCN), \gls{lstm}, and Temporal Fusion Transformers (TFT), complemented by \gls{shap}-based counterfactual explanations and attention mechanisms. Their research shows that interpretability significantly improves model reliability, particularly for detecting sleep disorders across heterogeneous populations. Similarly, Göktepe-Kavis et al.~\cite{Goktepe2024} emphasize the necessity of explainable methods within AI-driven sleep scoring and diagnostic frameworks. They highlight the critical role interpretability plays in mitigating algorithmic biases and fostering clinical trust. Furthermore, Kumar et al.~\cite{Kumar2023} stress the importance of interpretability for achieving standardized, generalizable AI models for sleep disorder detection. They argue that explainability is not only vital for user comprehension but also essential for regulatory compliance and ethical deployment in healthcare. Collectively, these studies highlight a growing consensus that explainability is indispensable in AI-driven sleep forecasting, particularly when transitioning predictive models from controlled laboratory settings to real-world clinical and personal healthcare applications.

\section{Dataset and pre-processing}

\subsection{Dataset}
This study was conducted in accordance with the Declaration of Helsinki and Dutch regulations governing research involving human participants. The Central Ethics Review Board for non-WMO studies (CTc UMCG), which oversees investigations outside the scope of the Medical Research Involving Human participants Act (WMO), reviewed and approved the study (study register number 18021). Written informed consent was obtained from each participant prior to enrollment.

Participants wore Garmin Vivosmart 5 devices for 10-15 weeks to record accelerations, heart rate, and blood oxygen saturation, with instructions to remove the device daily as recommended by the manufacturer. Participants were recruited from two field labs in northern Netherlands: a weekly walking group supervised by healthcare professionals, and a 10-week personal lifestyle intervention program featuring personalized nutrition, cooking workshops, sports sessions, and mindfulness lessons. Research team members attended weekly walking sessions and visited the intervention site biweekly to provide device assistance, while participants otherwise maintained their regular routines. Of the 17 initial participants, one withdrew after three weeks, leaving 16 female participants with a median age of 55.5 years (IQR: 16.25) and a median \gls{bmi} of 28.22 kg/m² (IQR: 10.74), seven of whom had prior smartwatch experience. All data are maintained in the secure, offline Virtual Research Workspace (VRW) administered by the University of Groningen. Prior to downloading and local analysis, data were preprocessed within the VRW in an offline environment to remove sensitive information. 


\begin{figure*}
    \centering
    \includegraphics[width=0.95\linewidth]{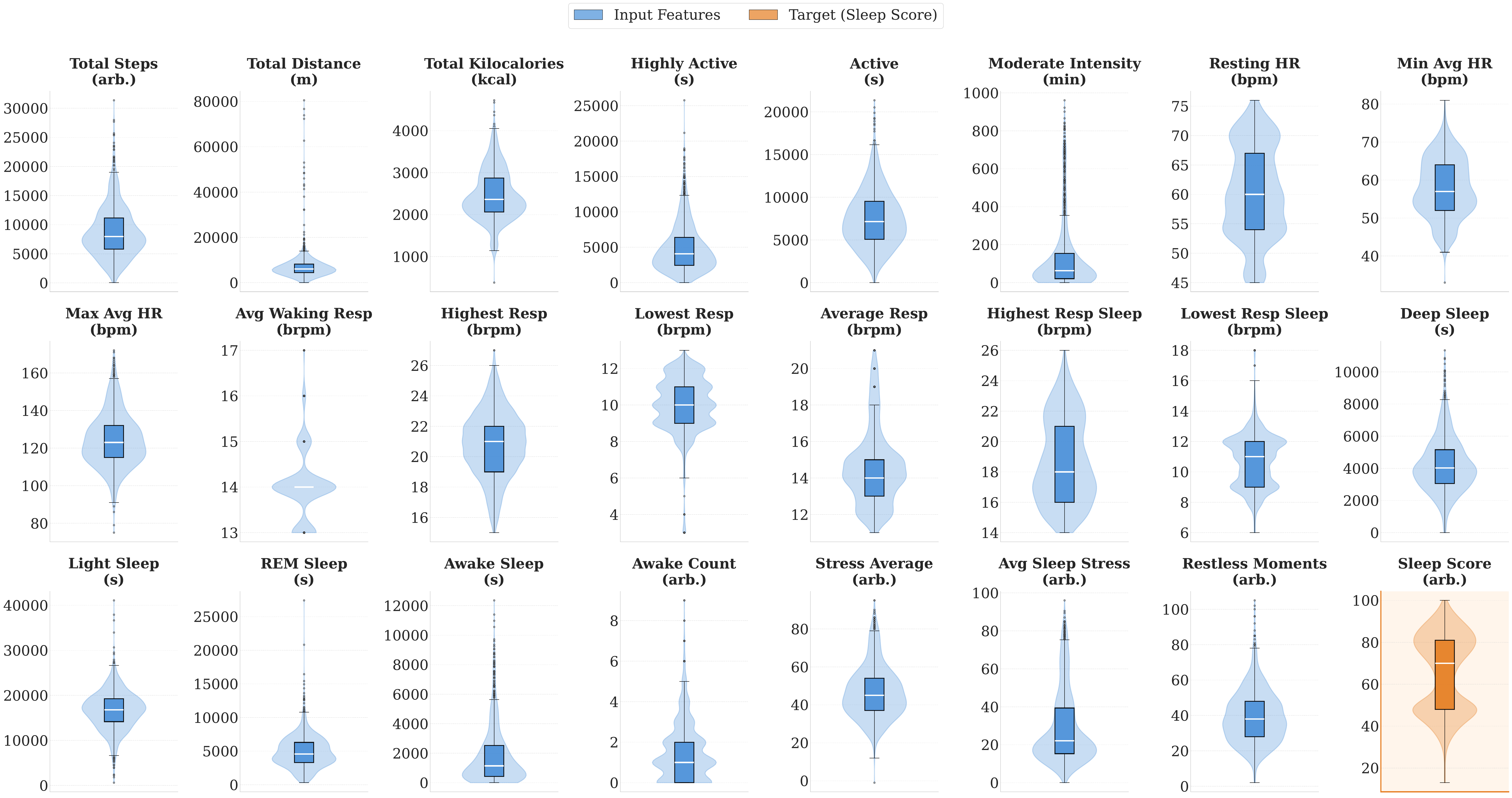}
    \caption{Distribution of 23 input features and the sleep score target using violin plots with embedded box plots. Features span five categories: activity, heart rate, respiration, sleep stages, and stress. Blue: input features; Orange: prediction target (sleep score).}
    \label{fig:box-plot-features}
\end{figure*}

\subsection{Features}
This study employs a single daily data point (i.e., sparse data) rather than original high-frequency measurements (one data point every 15 minutes) from Garmin, significantly reducing storage and computational demands. We removed the “Hydration” feature for all participants due to an excessive proportion of missing data because of manual input requirements, thereby ensuring a cleaner and more reliable data set. In addition, the first and last days of each participant’s recordings have been excluded due to insufficient device usage. A total of 24 features, which were directly exported from the Garmin Connect App, were used in this study, including total kilocalories, total steps, total distance, highly active seconds, active seconds, moderate-intensity minutes, resting heart rate, minimum average heart rate, maximum average heart rate, average waking respiration value, highest respiration value, lowest respiration value, stress average, deep sleep seconds, light sleep seconds, REM sleep seconds, awake sleep seconds, awake count, average sleep stress, restless moment count, lowest respiration, highest respiration, average respiration, and sleep scores as shown in Table~\ref{tab:acronyms} and Figure~\ref{fig:box-plot-features}. The predicting target, sleep score (0-100), is calculated using Firstbeat Analytic~\cite{kuula2021heart}.

\paragraph{Selection Methods}
We implement a comprehensive feature selection framework to identify the most relevant physiological and activity features for sleep quality prediction. Starting with 23 potential features encompassing activity metrics, heart rate, and respiration with cardio-respiratory metrics, respiration measurements, and sleep stage durations. We recognize that not all features contribute equally to prediction accuracy. Our framework allows dynamic feature selection through multiple methods, with the capability to reduce the feature set from 23 to 15 most informative features. By reducing from 24 to 15 features, we achieve approximately 32 percent dimensionality reduction, leading to faster training times, reduced overfitting risk, and improved model interpretability without sacrificing predictive performance. We design this selection process to occur after data preprocessing but before model training, ensuring that only meaningful features are used for learning sleep quality patterns.

First, we computed Pearson correlation coefficients between each feature and sleep quality scores across all time points. We rank features by absolute correlation values and select those exceeding a threshold of 0.05, ensuring we capture both positive and negative relationships. If fewer than 15 features meet this threshold, we include additional correlated features to maintain our target dimensionality. Second, our Recursive Feature Elimination (RFE) approach uses a Random Forest regressor as the base estimator to iteratively remove the least important features. We configure RFE to eliminate one feature at a time until reaching our target of 15 features, providing rankings that reflect non-linear feature importance. By reducing from 24 to 15 features, we achieve approximately 32 percent dimensionality reduction, leading to faster training times, reduced overfitting risk, and improved model interpretability without sacrificing predictive performance.

We further implement mutual information-based selection to capture non-linear dependencies between features and sleep quality. This method computes the mutual information between each feature and the target variable, selecting the top 15 features with the highest information content. This approach proves particularly valuable for identifying features with complex, non-monotonic relationships to sleep quality. Our most sophisticated approach is the ensemble feature selection method, which combines insights from all three previous methods. We run correlation, RFE, and mutual information selection with relaxed constraints (selecting top 15 features from each), then implement a voting mechanism where features receive votes based on their selection by each method. We select the 15 features with the highest vote counts, ensuring robust selection that is not biased by any single method's assumptions. 

We validate our feature selection approach through multiple mechanisms. We compare model performance with and without feature selection, consistently observing comparable or improved prediction accuracy with reduced feature sets. Our interpretability analysis using \gls{shap} values to confirm that selected features align with domain knowledge about sleep physiology. The consistency of selected features across different cross-validation folds demonstrates the stability of our selection process.

\subsection{Anomaly and missing data}
We implement a comprehensive multi-stage approach for handling anomalies and missing data in sleep quality prediction. The system first identifies missing data through two mechanisms: (i) invalid value detection, where sleep quality scores of -1 are replaced with NaN values, and (ii) zero value detection, where zero values in sleep quality scores are converted to NaN as they likely represent missing measurements. Missing and anomaly rates are summarized in Table~\ref{tab:acronyms}. 

Figure~\ref{fig:box-plot-features} reveals numerous outliers across 24 Garmin smartwatch features that are physiologically implausible for participants aged over 50. Notable anomalies include \texttt{totalSteps} exceeding 30,000 (typical active range: $<$10,000) and \texttt{highlyActiveSeconds} surpassing 20,000 (5.5 hours of high-intensity activity), both of which are highly unrealistic for this age demographic. The \texttt{awakeSleepSeconds} and \texttt{restlessMomentCount} features also show extreme values, suggesting sensor malfunction or improper device wear rather than genuine physiological measurements. These widespread anomalies across activity and sleep features necessitate robust anomaly detection before model training to prevent degraded sleep prediction performance and ensure reliable clinical insights. For anomaly detection, we employ two complementary approaches: a global statistical method using the interquartile range (IQR), which calculates Q1 (25th percentile) and Q3 (75th percentile) for each feature and defines anomalies as values outside [Q1 - 1×IQR, Q3 + 1×IQR], and a temporal deviation method that computes a 5-day rolling mean and identifies anomalies as values deviating more than 30 units from this mean. Following anomaly detection, we apply K-Nearest Neighbors (KNN) imputation with $k=3$ neighbors, which considers both cross-participant similarity and temporal patterns while preserving multivariate relationships between features. To address various types of noise in physiological time series data, we apply feature-specific smoothing methods based on physiological signal characteristics: exponential smoothing for heart rate (short-term trends), weighted moving average for activity (progressive patterns), adaptive smoothing for sleep stages (complex variability), Savitzky-Golay filtering for respiratory signals (feature preservation), and ensemble smoothing for the target sleep quality score (optimal robustness).



\subsection{Normalization}

We applied a MinMaxScaler from sklearn, scaling the feature data to a standard range (0 to 1) for train, validation, and test sets separately to avoid data leak. Each feature is normalized before being fed into the model. The predicted results are then denormalized to compare them with the true values and calculate the loss. 

\subsection{Sliding window}

To pre-process the time series data, we employ a sliding window approach with window size $w$ (3, 5, 7, 9, 11 days), stride $s$ (one day), and prediction window size $\delta$ (1, 3, 5, 7, 9 days), transforming each participant's dataset into a set of windowed instances for data augmentation as shown.

\begin{figure}
    \centering
    \includegraphics[width=1\linewidth]{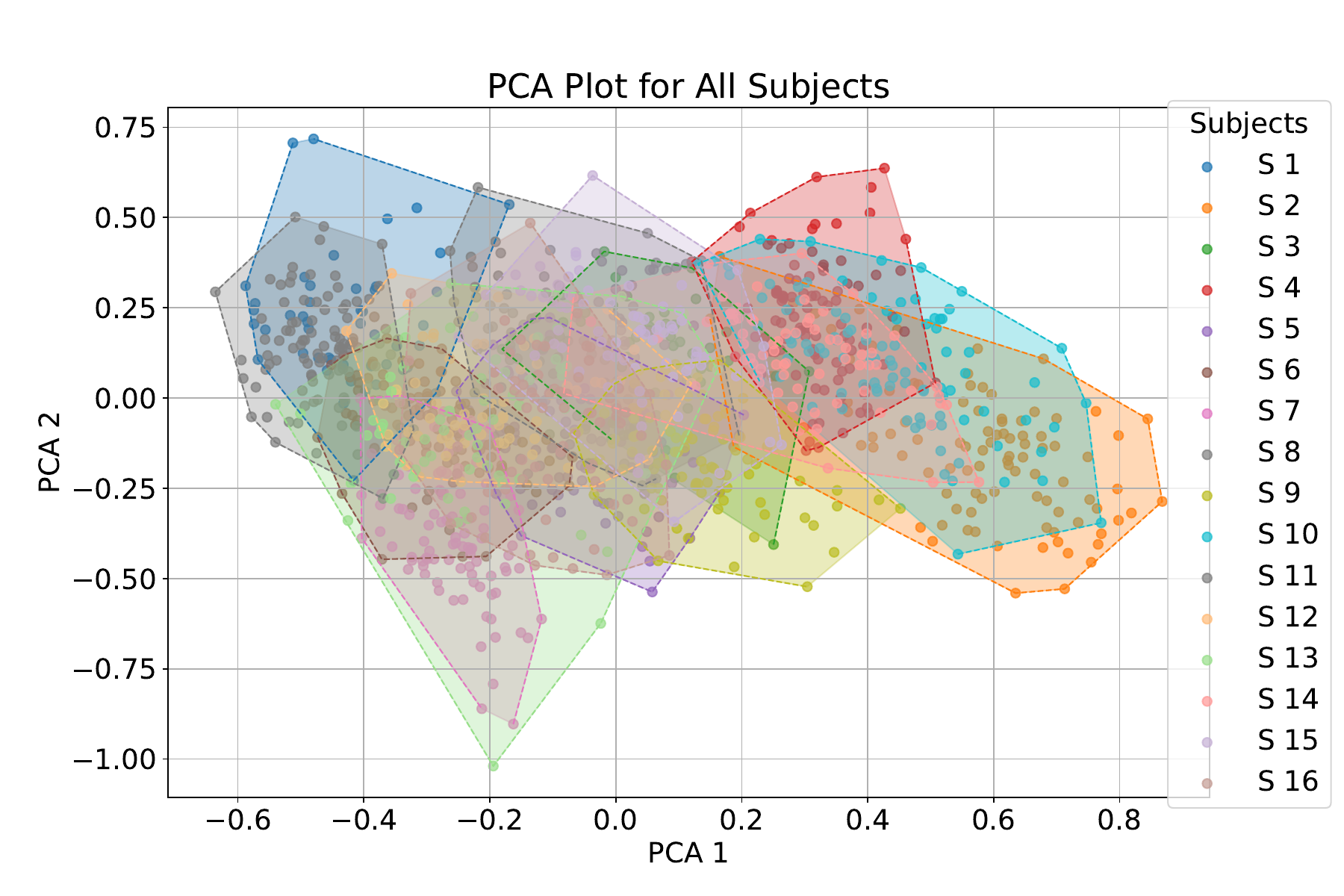}
    \caption{Domain shifts for features from different subjects by PCA.}
    \label{fig:domain-shift}    
\end{figure}

\subsection{Evaluation metrics}
\noindent
We use \gls{mse}, \gls{mae}, and Root Mean Squared Error (\gls{rmse}) as metrics in this work because they are frequently adopted in forecasting tasks, particularly in deep learning contexts, due to their interpretability and computational simplicity. They are defined as:

\begin{equation}
\text{\gls{mse}} = \frac{1}{n} \sum_{i=1}^n (y_i - \hat{y}_i)^2
\end{equation}

\begin{equation}
\text{\gls{mae}} = \frac{1}{n} \sum_{i=1}^n |y_i - \hat{y}_i|
\end{equation}

\begin{equation}
\text{\gls{rmse}} = \sqrt{\frac{1}{n} \sum_{i=1}^n (y_i - \hat{y}_i)^2}
\end{equation}

where \(y_i\) denotes the normalized sleep value on day \(i\), \(\hat{y}_i\) represents the corresponding predicted normalized value, and \(n\) is the total number of days.

\subsection{Leave-one-out cross-validation}
In many machine learning problems involving human participants, data collected from different individuals may vary significantly due to physiological, behavioral, or demographic factors. One standard approach for evaluating the robustness of a model across different individuals is \emph{\gls{loocv}}. This approach ensures that for each participant, the model is evaluated on data that remain completely unseen during training, providing a realistic estimate of cross-participant generalization performance. There are 16 distinct participants in this work, each associated with a dataset:
\begin{equation}
\mathcal{S}_p = \{(\mathbf{f}_{p,k}, q_{p,k}) \mid k = 1, 2, \ldots, T_p\},
\end{equation}
where:
\begin{itemize}
    \item $\mathbf{f}_{p,k} \in \mathcal{F}$ is the feature vector of the $k$-th sample from participant $p$,
    \item $q_{p,k} \in \mathcal{Q}$ is the sleep quality score corresponding to $\mathbf{f}_{p,k}$,
\end{itemize}
and $T_p$ is the total number of samples from participant $p$.
Hence, the entire dataset by 16 participants in this work can be written as
\begin{equation}
\mathcal{D} = \bigcup_{s=1}^{16} \mathcal{D}_s.
\end{equation}
In this work, we split the data by participant. Concretely, for each participant $s \in \{1, 2, \ldots, 16\}$:
Practitioners split the \emph{training data} (from the $N$ subjects) into a separate training set ($N - 2$ participants), a validation set (one participant), and a test set (one participant). For instance,
\begin{equation}
\text{Train set} = \mathcal{D}_a, \quad
\text{Validation set} = \mathcal{D}_b, \quad
\text{Test set} = \mathcal{D}_s,
\end{equation}
where $a \neq s$, $b \neq s$, and $a \neq b$. Hyperparameters are tuned on $\mathcal{D}_b$ (validation set), while ultimately evaluating performance on the unseen participant $\mathcal{D}_s$ (test set).

\begin{table*}[htbp]
\centering
\caption{Performance comparison of \gls{rmse} for various models and input and prediction window sizes}
\label{tab:model_comparison_complete}
\scriptsize
\begin{tabular}{l|c|c|c|c|c||c|c|c|c|c||c|c|c|c|c}
\toprule[1.5pt]
\multirow{2}{*}{\textbf{Model}} & \multicolumn{5}{c||}{\textbf{Input = 3 days}} & \multicolumn{5}{c||}{\textbf{Input = 5 days}} & \multicolumn{5}{c}{\textbf{Input = 7 days}} \\
\cline{2-16}
 & 1 & 3 & 5 & 7 & 9 & 1 & 3 & 5 & 7 & 9 & 1 & 3 & 5 & 7 & 9 \\
\midrule
\gls{lstm}~\cite{hochreiter1997long} & 0.300 & 0.304 & 0.298 & 0.316 & 0.299 & 0.312 & 0.299 & 0.305 & 0.297 & 0.310 & 0.332 & 0.308 & 0.336 & 0.335 & 0.321 \\
Informer~\cite{zhou2020informer} & 0.290 & 0.290 & 0.290 & 0.293 & 0.290 & 0.293 & 0.297 & 0.294 & 0.292 & 0.298 & 0.296 & 0.294 & 0.294 & 0.293 & 0.293 \\
PatchTST~\cite{nie2022time} & 0.292 & 0.294 & 0.296 & 0.297 & 0.299 & 0.285 & 0.294 & 0.301 & 0.299 & 0.299 & 0.299 & 0.291 & 0.308 & 0.303 & 0.317 \\
TimesNet~\cite{wu2022timesnet} & 0.283 & 0.294 & 0.297 & 0.329 & 0.322 & 0.279 & 0.293 & 0.305 & 0.303 & 0.330 & 0.294 & 0.292 & 0.294 & 0.307 & 0.312 \\
\textbf{Ours} & \textbf{0.216} & 0.257 & 0.272 & 0.316 & 0.317 & 0.231 & 0.234 & 0.229 & 0.317 & 0.318 & 0.279 & 0.286 & 0.249 & 0.313 & 0.320 \\
\bottomrule[1.5pt]
\end{tabular}

\vspace{10pt} 

\begin{tabular}{l|c|c|c|c|c||c|c|c|c|c}
\toprule[1.5pt]
\multirow{2}{*}{\textbf{Model}} & \multicolumn{5}{c||}{\textbf{Input = 9 days}} & \multicolumn{5}{c}{\textbf{Input = 11 days}} \\
\cline{2-11}
 & 1 & 3 & 5 & 7 & 9 & 1 & 3 & 5 & 7 & 9 \\
\midrule
\gls{lstm}~\cite{hochreiter1997long} & 0.300 & 0.305 & 0.333 & 0.320 & 0.344 & 0.289 & 0.308 & 0.304 & 0.315 & 0.312 \\
Informer~\cite{zhou2020informer} & 0.292 & 0.292 & 0.295 & 0.294 & 0.291 & 0.290 & 0.292 & 0.294 & 0.292 & 0.291 \\
PatchTST~\cite{nie2022time} & 0.279 & 0.302 & 0.295 & 0.321 & 0.328 & 0.307 & 0.310 & 0.317 & 0.310 & 0.308 \\
TimesNet~\cite{wu2022timesnet} & 0.291 & 0.299 & 0.313 & 0.319 & 0.314 & 0.285 & 0.303 & 0.309 & 0.315 & 0.322\\
\textbf{Ours} & 0.307 & 0.311 & 0.267 & 0.306 & 0.317 & 0.265 & 0.287 & 0.297 & 0.321 & 0.313 \\
\bottomrule[1.5pt]
\end{tabular}
\end{table*}

After performing this procedure for each participant $i$, we obtain a set of metrics on the $i$-th fold: \gls{mae} $\{MAE_1, \ldots, MAE_N\}$, MSE $\{MSE_1, \ldots, MSE_N\}$, and \gls{rmse} $\{RMSE_1, \ldots, RMSE_N\}$. \gls{loocv} is a rigorous strategy for evaluating participant-based machine learning tasks. By training on $N-2$ participants, validating on another participant, and testing on the remaining participant, we obtain an unbiased estimate of cross-participant generalization. 

\begin{algorithm}[H]
\caption{Our Adaptive Multi-scale Temporal Framework}
\label{alg:our-model}
\footnotesize  
\textbf{Input:} $\mathcal{D}_{\text{train}}, \mathcal{D}_{\text{val}}, \mathcal{D}_{\text{test}}$, Epochs $E$, Domain weight $\alpha$
  \textbf{Output:} Trained model $\mathcal{M}$
\begin{algorithmic}[1]
\STATE \textbf{Phase 1: Training with Domain Adaptation}
\STATE Initialize $\mathcal{M}$ (Multi-scale \gls{cnn} + BiLSTM + Self-Attention + Domain Classifier)
\STATE Initialize Adam optimizer with $\text{lr} = 10^{-3}$
\FOR{epoch $= 1$ to $E$}
  \FOR{batch $(X, y, d) \in \mathcal{D}_{\text{train}}$}
      \STATE $\hat{y}, \hat{d} \leftarrow \mathcal{M}(X, \text{return\_domain}=\text{True})$
      \STATE $\mathcal{L} \leftarrow \text{MSE}(\hat{y}, y) + \alpha \cdot \text{CrossEntropy}(\hat{d}, d)$
      \STATE Update $\mathcal{M}$ via backpropagation
  \ENDFOR
  \STATE Evaluate on $\mathcal{D}_{\text{val}}$ and apply Early Stopping
\ENDFOR
\STATE \textbf{Phase 2: Test-Time Adaptation}
\STATE Initialize \gls{tta} optimizer with $\text{lr}_{\text{tta}} = 10^{-4}$
\FOR{epoch $= 1$ to $E_{\text{tta}}$}
  \FOR{batch $(X_{\text{test}}, -, -) \in \mathcal{D}_{\text{test}}$}
      \STATE Generate $X_{\text{aug1}}, X_{\text{aug2}}$ with noise augmentation
      \STATE $\hat{y}_{\text{orig}} \leftarrow \mathcal{M}(X_{\text{test}})$, $\hat{y}_{\text{aug1}} \leftarrow \mathcal{M}(X_{\text{aug1}})$, $\hat{y}_{\text{aug2}} \leftarrow \mathcal{M}(X_{\text{aug2}})$
      \STATE $\mathcal{L}_{\text{tta}} \leftarrow \frac{1}{3}\sum \text{MSE}(\hat{y}_i, \hat{y}_j)$ for all pairs
      \STATE Update $\mathcal{M}$ via backpropagation
  \ENDFOR
\ENDFOR
\STATE \textbf{Phase 3: Final Inference}
\STATE $\mathcal{M}.\text{eval}()$
\FOR{batch $(X_{\text{test}}, y_{\text{test}}, -) \in \mathcal{D}_{\text{test}}$}
  \STATE $\hat{y}_{\text{final}} \leftarrow \mathcal{M}(X_{\text{test}})$
\ENDFOR
\STATE Compute \gls{rmse}, \gls{mae}, MSE
\end{algorithmic}
\end{algorithm}

\section{Methodology}

\subsection{Models}

\begin{figure*}
    \centering
    \includegraphics[width=1\linewidth]{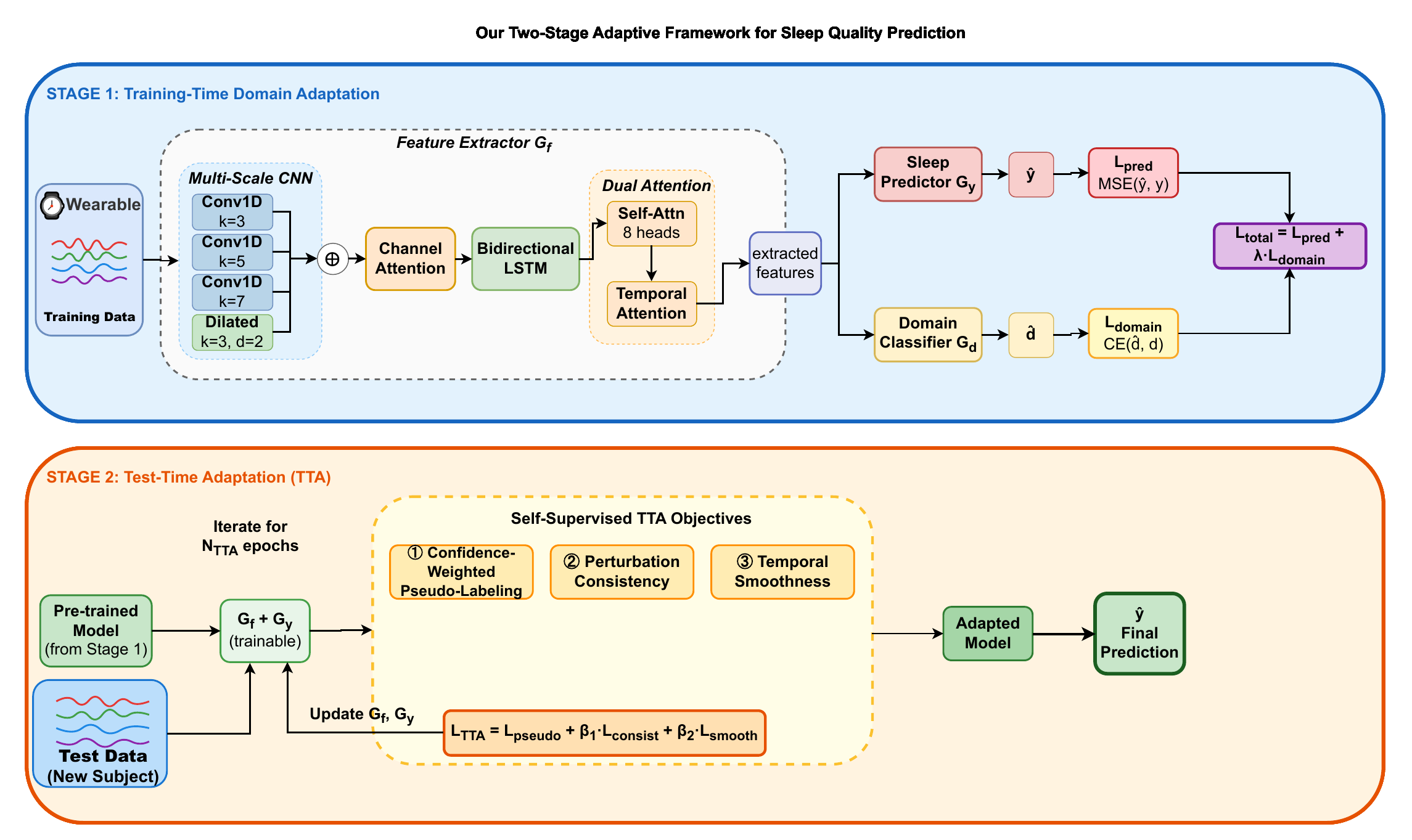}
    \caption{Flowchart of our adaptive spatial-temporal model with optional two-stage domain adaptation}
    \label{fig:model-flowchart}
\end{figure*}

We propose a spatial-temporal model with attention mechanisms as our primary architecture for sleep quality prediction, as shown in the flowchart of Figure~\ref{fig:model-flowchart}. This model was specifically designed to address the unique challenges of multi-modal physiological time series data, which exhibits both local patterns (e.g., short-term heart rate variations) and long-range temporal dependencies. 

Our architecture integrates multiple state-of-the-art components: a multi-scale \gls{cnn} module with parallel convolutional layers and dilated convolutions, a channel attention mechanism for automatic feature selection, a bidirectional temporal encoder layer for capturing temporal dependencies in both directions, and dual attention mechanisms including self-attention and temporal attention for identifying crucial time points.

\subsubsection{Multi-Scale Convolutional Module}
Sleep-related patterns manifest at multiple temporal scales. To capture this inherent multi-scale structure, we employ parallel convolutional branches with kernel sizes of 3, 5, and 7, corresponding to short-term (1-day), medium-term (2-day), and longer-term (3-day) patterns in our daily-aggregated data. We additionally incorporate a dilated convolution branch (kernel size 3, dilation rate 2) that extends the effective receptive field to capture weekly periodicity without increasing parameter count. The outputs from all branches are concatenated to form a rich multi-scale representation. This design differs fundamentally from standard single-scale CNNs: rather than relying on deep stacking to gradually expand the receptive field, our parallel architecture explicitly extracts patterns at physiologically meaningful time scales from the first layer, enabling more efficient and interpretable feature learning.



\subsubsection{Temporal modules with Dual Attention}

While the CNN module excels at extracting local patterns, sleep quality prediction requires understanding longer-range temporal dependencies—for example, how accumulated sleep quality over the past week affects tonight's sleep. We employ a bidirectional LSTM that processes the CNN features in both forward and backward directions, enabling the model to leverage both historical context and future patterns within the input window.

Building upon the LSTM representations, we introduce a dual attention mechanism that serves two complementary purposes. The first component, an 8-head self-attention layer, models pairwise interactions between all time steps, capturing non-local dependencies such as weekend-weekday contrasts or the relationship between exercise days and subsequent recovery nights. The second component, a temporal attention layer, computes a learned weighted combination over all time steps to produce the final representation. Unlike self-attention, which models relationships, temporal attention explicitly identifies which specific days within the input window are most informative for predicting current sleep quality—effectively learning that recent nights typically matter more than distant ones, while still allowing the model to attend to earlier anomalous events when relevant. The combination of self-attention and temporal attention provides richer temporal modeling than either mechanism alone: self-attention captures \textit{how} different time points relate to each other, while temporal attention determines \textit{which} time points matter most for the final prediction.

\subsection{Domain Adaptation}
Figure~\ref{fig:domain-shift} illustrates a significant domain shift in our dataset using PCA visualization across all subjects. Each participant's data forms distinct, minimally overlapping clusters in the reduced feature space, with subjects distributed across different regions—for example, Subject 3 occupies the lower-left quadrant while Subject 4 clusters in the upper-right. The convex hulls emphasize this separation, with some subjects (S1, S8, S10) showing compact clusters and others (S3, S7, S14) exhibiting dispersed patterns. This substantial inter-subject variability in physiological and behavioral patterns demonstrates that traditional assumptions fail for this data, necessitating domain adaptation techniques to develop robust sleep prediction models that generalize across diverse individual profiles.

This distributional shift can degrade the model's performance on unseen participants. $\mathcal{D}_a = \mathcal{D}_{\text{train}}$ represent the \emph{source domain} and $\mathcal{D}_t = \mathcal{D}_q$ denote the \emph{target domain} for a given fold $(i, q)$. The joint distributions of the source and target domains are:
\[
P_s(\mathbf{x}, \mathbf{y}) \; \neq \; P_t(\mathbf{x}, \mathbf{y}),
\]
where $P_s$ and $P_t$ are the probability distributions over $\mathcal{X} \times \mathcal{Y}$ for the source and target domains, respectively.


Training on $\mathcal{D}_p$, validating on $\mathcal{D}_q$, and testing on $\mathcal{D}_i$ can be interpreted as adapting from a \emph{source domain} to a \emph{target domain}. This viewpoint motivates the application of \emph{domain adaptation} methods, which aim to reduce distributional discrepancies between training participants and a new target participant.


\paragraph{Our Domain Adaptation Strategy}
We designed our framework to ensure that models trained on one group of participants can effectively generalize to new, unseen participants. We implement a two-stage domain adaptation approach to address the critical challenge of cross-participant generalization in sleep quality prediction, as shown in the pseudo code from Algorithm~\ref{alg:our-model}.  Our adaptation strategy operates in two distinct stages: training-time domain adaptation and \gls{tta}. We provide flexible configuration options allowing these stages to be used independently or in combination through four modes: 'none', 'train-only', 'test-only', or 'both', enabling systematic evaluation of each component's contribution.

\paragraph{Stage 1: Training-Time Domain Adaptation}
In the first stage, we implement domain adaptation during model training through adversarial learning. We equip each model architecture with a domain classifier component designed to distinguish between different participants (domains). Our training process employs a minimax game where the feature extractor learns representations that are simultaneously predictive of sleep quality and invariant across participants. We formulate the total loss as a combination of the primary sleep quality prediction loss and a domain classification loss, controlled by a weighting parameter alpha. Through gradient reversal, we train the domain classifier to identify the source participant while simultaneously training the feature extractor to produce representations that confuse the domain classifier. This adversarial training mechanism encourages our models to learn participant-invariant features that enhance generalization to new individuals.
\paragraph{Stage 2: Test-Time Adaptation (TTA)}
We introduce the second adaptation stage during inference when our model encounters data from previously unseen participants. We implement three complementary \gls{tta} strategies. First, our entropy-based \gls{tta} leverages the model's prediction confidence for self-supervised adaptation. We identify high-confidence predictions (using a threshold of 0.9) and employ them as pseudo-labels to fine-tune the model on the test participant's unlabeled data. Second, our consistency-based \gls{tta} enforces prediction stability under input perturbations. We update the model to minimize discrepancies between predictions on original inputs and their augmented versions (with controlled noise levels of 0.01 and 0.02). Third, we develop a temporal consistency \gls{tta} that exploits the sequential nature of sleep data by enforcing smooth predictions across time and maintaining consistency between temporally shifted sequences.

\paragraph{Implementation Design and Flexibility}
We designed our framework with flexibility as a core principle, allowing systematic investigation of each adaptation component's effectiveness. We enable four operational modes: setting mode to 'none' provides baseline performance without adaptation, 'train-only' isolates the effect of adversarial training during the learning phase, 'test-only' evaluates the impact of test-time adaptation alone, and 'both' combines both stages for comprehensive adaptation. We configure training-time adaptation with a domain loss weight of 1.0 and implement gradient reversal for effective adversarial feature learning. For test-time adaptation, we employ 20 adaptation epochs with a learning rate of $1 \times 10^{-4}$, with the specific strategy determined by the \gls{tta}-method parameter.

\paragraph{Rationale and Empirical Benefits}
The training-time adaptation ensures our models learn robust and generalizable features across the training population, mitigating overfitting to individual participants' patterns. The test-time adaptation enables personalization when deploying the model to new participants, allowing dynamic adjustment to individual physiological characteristics without requiring additional labeled data. Our experimental results validate this design through systematic comparison across adaptation modes. The pseudo code is shown above in the info-graphic Algorithm~\ref{alg:our-model}.

\subsection{Baseline Models for Comparative Analysis}
To evaluate our proposed architecture, we implemented several baseline models representing different modeling paradigms. The \gls{lstm} Model uses stacked \gls{lstm} layers without convolutional preprocessing, evaluating the importance of explicit local feature extraction. We also compared against recent transformer-based architectures that have shown promise in time series forecasting, including PatchTST~\cite{nie2022time}, Informer~\cite{zhou2020informer}, and timesnet~\cite{wu2022timesnet}. 

\subsection{Explainable AI for sleep forecast}


We implement \gls{shap} analysis~\cite{lundberg2017unified} using the model-agnostic KernelExplainer to interpret our deep learning models. To handle the three-dimensional input data (samples × time steps × features), we apply temporal aggregation, using 50 background samples and 100 test samples for computational efficiency. We calculate \gls{shap} values through the standard Shapley value formula~\cite{lundberg2017unified}. This approach provides both feature rankings and directional insights, providing an understanding not only of which factors matter but also of how they influence sleep quality. The framework supports personalized explanations through individual values while maintaining computational tractability through temporal aggregation and strategic sampling.

\subsection{Training Procedure}
Our training methods consist of two optional phases. In Phase 1, the model could be trained with domain adaptation using a combined loss function $\mathcal{L} = \mathcal{L}_{\text{main}} + \alpha \cdot \mathcal{L}_{\text{domain}}$, where $\mathcal{L}_{\text{main}} = \text{\gls{rmse}}(\hat{y}, y)$ represents the sleep quality prediction loss and $\mathcal{L}_{\text{domain}} = \text{CrossEntropy}(\hat{d}, d)$ enforces domain-invariant feature learning across participants, with the domain weight $\alpha$ optimized during hyperparameter tuning. The training employs Adam optimizer with learning rate $10^{-3}$, which incorporates early stopping with smoothed validation loss (patience=30, min\_delta=0.0001), and applies comprehensive data preprocessing including ensemble smoothing methods and automated feature selection to 15 most relevant features from the original 23 Garmin metrics. In Phase 2: we implement \gls{tta} on the held-out test participant using consistency regularization with augmented inputs, minimizing $\mathcal{L}_{\text{TTA}} = \frac{1}{3}\sum_{i,j} \text{\gls{rmse}}(\hat{y}_i, \hat{y}_j)$ across original and noise-augmented predictions with learning rate $10^{-4}$ for 10 epochs. Phase 3 conducts final inference on the adapted model, and this entire procedure is repeated for each participant in the \gls{loocv} framework, ensuring robust evaluation across individual differences, while the domain adaptation components help the model generalize across different participants' physiological patterns and behavioral variations in sleep data.

\subsection{Hyperparameter Optimization}
All models in this work were optimized using Optuna's Tree-structured Parzen Estimator (TPE) sampling within a \gls{loocv} framework to ensure robust generalization across participants. The search space included architectural parameters: number of convolutional layers $\in \{1, 2\}$, \gls{lstm} layers $\in \{1, 2, 3\}$, \gls{cnn} hidden dimensions $\in \{16, 32, 64\}$, \gls{lstm} hidden dimensions $\in \{64, 128, 256\}$, and regularization parameters: \gls{cnn} dropout rate $\in [0.1, 0.5]$, \gls{lstm} dropout rate $\in [0.1, 0.5]$, batch normalization toggle $\in \{\text{True}, \text{False}\}$, batch size $\in \{8, 16, 32\}$, and domain adaptation weight $\alpha \in [0.0, 1.0]$. The optimization objective minimized the average \gls{rmse} across all \gls{loocv} folds: $\theta^* = \arg\min_{\theta \in \Theta} \frac{1}{N} \sum_{i=1}^{N} \text{\gls{rmse}}(\mathcal{M}_\theta, S_i)$, where $\mathcal{M}_\theta$ represents the model parameterized by $\theta$, and $S_i$ denotes the $i$-th held-out participant. Early stopping with smoothed validation loss (smoothing factor $\beta = 0.1$) was applied with patience of 30 epochs and minimum improvement threshold of 0.0001 to prevent overfitting during the hyperparameter search process. 

\subsection{Hardware and Software Setup}
\label{sec:hardware}

Experiments were conducted on two hardware platforms: a local workstation (Intel i7-13700K, NVIDIA RTX 4090, 128 GB RAM) for test and development, and the Hábrók HPC cluster\footnote{\url{https://wiki.hpc.rug.nl/habrok/introduction/cluster_description}} for large-scale computations like hyperparameter fine-tuning. The cluster comprises 190 heterogeneous nodes including standard compute nodes (119×128 AMD EPYC cores), high-memory nodes (4×80 Intel Xeon cores, 4 TB RAM), and GPU nodes featuring NVIDIA A100, V100, L40s, and H100 accelerators. All models were implemented in PyTorch for cross-platform consistency.

\section{Results and Discussion}

Our approach captures both local feature interactions and longer-term temporal dependencies, while a two-stage adaptation provides adaptation across different participants. Our model offers several critical advantages for sleep quality prediction. First, the multi-scale feature extraction capability allows the model to simultaneously capture patterns at different temporal resolutions—from minute-level heart rate fluctuations to hour-level sleep stage progressions. Second, the attention mechanisms provide interpretability by highlighting which features and time points most influence predictions, crucial for clinical applications. Third, the bidirectional processing leverages both historical context and future patterns within the input window, particularly valuable for understanding sleep preparation and recovery phases. Fourth, residual connections between \gls{cnn} and output layers ensure stable gradient flow during training while preserving fine-grained temporal information. Finally, the integrated domain adaptation components enable robust cross-participant generalization, addressing the significant inter-individual variability in sleep patterns. We observe consistent improvements across all metrics (\gls{mae}, MSE, and \gls{rmse}) when employing domain adaptation. The most substantial gains are achieved when both adaptation stages are active (mode='both'), yielding an \gls{rmse} of 0.216 compared to 0.245 for training-only adaptation and 0.242 for test-only adaptation. This finding confirms our hypothesis that combining population-level robustness (from training-time adaptation) with individual-level personalization (from test-time adaptation) provides optimal performance for cross-participant sleep quality prediction. Our approach is particularly valuable in real-world healthcare applications where physiological variability is substantial and obtaining labeled data from every new patient is often impractical. 

\subsection{Evaluation with Different Window Sizes}
We tested five input window sizes (3, 5, 7, 9, and 11 days) in combination with five predicting window sizes (1, 3, 5, 7, and 9 days) on both baseline models and our proposed adaptive spatial and temporal model. Table~\ref{tab:model_comparison_complete} summarizes the results. From the results, it can be observed that all models, including our proposed approach, generally achieve their lowest \gls{rmse} when the prediction window size is one day. Longer prediction windows are also possible with the increase in errors. 
Our model achieves the best performance compared to other base models and state-of-the-art time series forecasting models. 

\begin{figure*}
    \centering
    \includegraphics[width=1\linewidth]{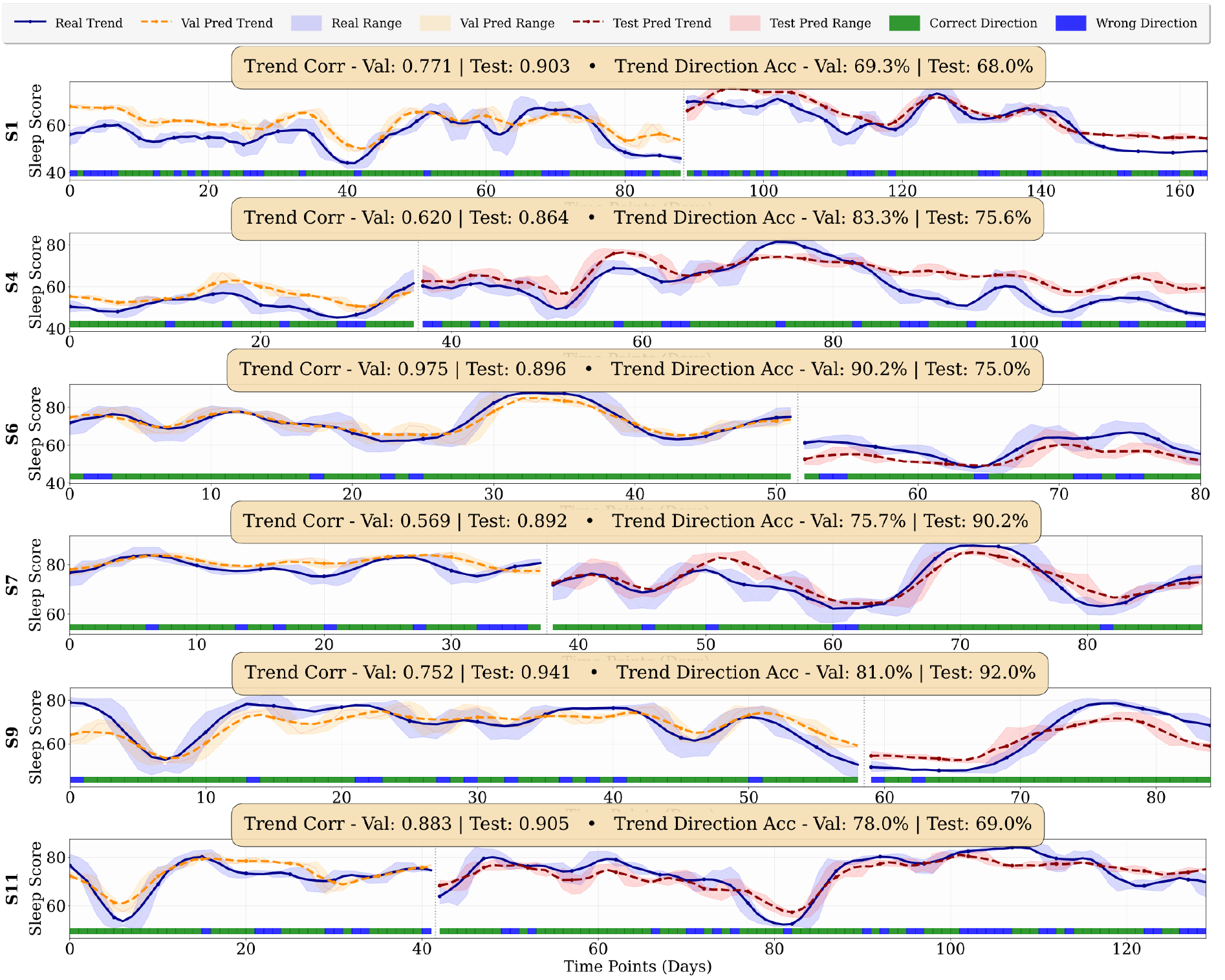}
    \caption{True values vs predictions in validation and test set for subjects. Sleep quality prediction results across subjects representing typical patterns showing real trends (solid blue lines) compared to validation predictions (orange dashed) and test predictions (red dashed) with uncertainty bands (shaded regions). Color-coded bars indicate trend direction accuracy (green = correct, blue = incorrect).}
    \label{fig:true-pre}
\end{figure*}

\subsection{Temporal Prediction Analysis and Trend Accuracy}
Figure~\ref{fig:true-pre} presents a comparison between ground-truth sleep scores and model predictions across both validation and test sets. To maintain clarity and avoid redundancy, we display representative participants who demonstrate the range of model performance patterns observed across various participants. The visualizations reveal that our model not only accurately predicts overall sleep quality levels but also effectively captures day-to-day variations and long-term sleep pattern changes for most participants. 

To comprehensively evaluate model performance, we simultaneously display raw predictions and smoothed trend lines. Raw data is shown with faded colors, while trend lines are highlighted using darker colors and thicker lines. Validation and test sets are distinguished through different color schemes: blue-orange for the validation set and blue-red for the test set. By calculating rolling standard deviations, we add confidence bands to each prediction using semi-transparent fills to represent prediction uncertainty ranges. This visualization method intuitively demonstrates the stability of model predictions. We separately calculate Pearson correlation coefficients for validation and test sets to evaluate the model's ability to capture sleep quality trends. Beyond traditional regression metrics, we introduce trend direction accuracy to evaluate the model's ability to predict sleep quality change directions. 

Through trend correlation assessment, we can verify whether the model captures long-term sleep quality changes, which is crucial for monitoring chronic sleep disorders. Trend direction accuracy directly reflects the model's ability to predict sleep quality improvement or deterioration, providing important guidance for timely interventions. By displaying prediction range bands, clinicians can understand prediction confidence levels, facilitating more cautious decision-making. Our visualization method uses color coding (green for correct direction predictions, blue for incorrect) to display model performance at each time point, enabling quick identification of model strengths and weaknesses.

Using rolling window methods to calculate trends and standard deviations improves evaluation robustness and reduces outlier effects. The visualization in Figure~\ref{fig:true-pre} reveals a strong alignment between predicted and actual sleep scores across most participants. Several key observations emerge:

\textbf{Trend Correlation Performance}: The trend correlation values demonstrate robust model performance, test set correlations spanning 0.864 (participant 4) to 0.941(participant 9). The consistently high correlation values (most exceeding 0.88) indicate that our model effectively captures the underlying sleep quality patterns across diverse individuals.

\textbf{Trend Direction Accuracy}: The trend direction accuracy metric shows performance, with test
set accuracies ranging from 68.0\% (participant 1) to 92.0\% (participant 7). These high accuracy rates demonstrate the model's ability to correctly predict whether sleep quality will improve or deteriorate, which is crucial for clinical intervention timing.

\subsubsection{Individual Participant Analysis}

The participant-specific patterns reveal interesting variations in model performance. Participants (4, 7, 9) exhibit exceptional model fit with both high trend correlations ($>$0.864) and direction accuracies ($>$75.6\%). The visual alignment between predicted and actual values is particularly strong, with the model successfully capturing both short-term fluctuations and long-term trends. While still achieving respectable performance metrics, participants (7, 9) show slightly more prediction variance. Participant 7, for instance, demonstrates good test set performance (90.2\% direction accuracy) despite having the lowest validation correlation (0.569), suggesting successful model adaptation during testing. Participants with highly variable sleep patterns (e.g., participant 11's dramatic fluctuations around day 100) present greater prediction challenges, yet the model maintains reasonable accuracy by correctly capturing the general trend direction even when exact values differ.

\subsubsection{Temporal Dynamics}

The temporal analysis reveals several important characteristics. The gray vertical line separating validation and test sets shows generally smooth transitions in prediction quality, indicating robust model generalization. The absence of significant performance degradation in test sets validates our training methodology. The shaded regions representing prediction ranges effectively bound the actual values in most cases, with wider uncertainty bands appearing during periods of higher sleep variability. This adaptive uncertainty quantification provides valuable confidence information for clinical decision-making. The color-coded bars at the bottom of each plot (green for correct direction, blue for incorrect) provide immediate visual feedback on prediction reliability. Notably, clusters of correct predictions often coincide with stable sleep patterns, while incorrect predictions tend to occur during rapid transitions.

\subsubsection{Clinical Relevance}
The high trend direction accuracies across participants have clinical implications: With high direction accuracies, clinicians can confidently use model predictions to identify when a patient's sleep quality is likely to deteriorate, enabling proactive interventions. The variation in model performance across participants (e.g., participant 9's 0.941 test correlation vs. participant 6's 0.896) suggests that some individuals have more predictable sleep patterns, allowing for customized monitoring frequencies. Additionally, the strong trend correlations indicate that the model captures meaningful sleep quality evolution rather than just noise, further supporting its clinical utility for personalized sleep health management.

\subsubsection{Model Robustness}
The analysis demonstrates several aspects of model robustness. Despite individual sleep pattern variations, the model maintains high performance metrics across all participants. The model performs consistently across extended time periods (up to 170 days for some participants), suggesting good long-term reliability. The model successfully adapts to different sleep pattern types, from highly regular (participant 6) to more variable (participant 11), demonstrating flexible feature learning.

These results collectively validate our model's clinical utility for continuous sleep quality prediction, providing both accurate point estimates and reliable trend information essential for effective sleep disorder management.

\begin{figure}
    \centering
    \includegraphics[width=0.9\linewidth]{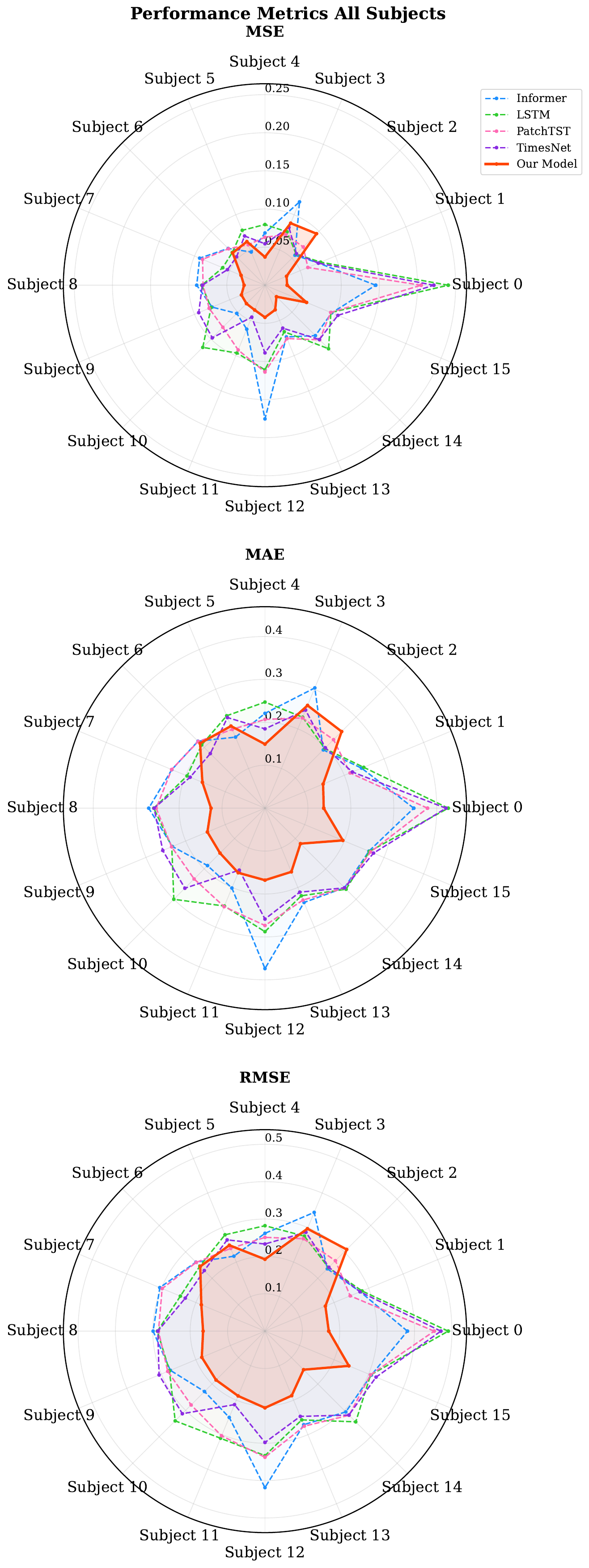}
    \caption{Comparison of ALL participants' \gls{mae}, MSE, and \gls{rmse} by different models in terms of three-day input and one-day prediction window size. }
    \label{fig:radar-plot}
\end{figure}
\subsection{Cross-participant Performance Comparison}
Figure~\ref{fig:radar-plot} presents a comprehensive comparison of model performance across all participants using three key metrics: Mean Squared Error (MSE), Mean Absolute Error (\gls{mae}), and Root Mean Squared Error (\gls{rmse}) in terms of three-day input and one-day prediction window size. The radar plots enable simultaneous visualization of individual participant performance and overall model behavior patterns.

The shorter radial distances for our model indicate substantially lower error compared to the baselines in almost every case. This consistency suggests that our spatial and temporal model with two-stage domain adaptation techniques enables our approach to robustly handle participant-specific variability and produce accurate sleep forecasts.

Overall, these findings validate the effectiveness of the proposed method. The model not only excels with shorter predicting windows (one day) but also sustains competitive accuracy for extended horizons. Such versatility could be valuable for real-world applications where sleep monitoring and prediction over multiple days can inform better health interventions and personalized recommendations.

\subsubsection{Model Performance Patterns}

The radar plots reveal several important patterns in model performance:

\textbf{Overall Performance Hierarchy}: Across all three metrics, a consistent performance ranking emerges, with our proposed model demonstrating superior performance (smaller error areas) compared to baseline methods. Notable performance variations exist across participants: participants 1, 5, 6, and 13 show relatively low error values across all models, suggesting these participants have more predictable sleep patterns.
It exhibits higher error values, particularly pronounced in the MSE and \gls{rmse} metrics, indicating greater prediction challenges for participants 0, 10, and 12. Participant 7 demonstrates exceptional model performance with minimal errors across all metrics, suggesting highly regular sleep patterns.

\subsubsection{Metric-Specific Observations} The MSE plot (top panel) shows the error values ranging from near 0 to 0.25. The quadratic nature of MSE amplifies the performance gaps, making it particularly useful for identifying participants where models struggle. Our model maintains the smallest area coverage, indicating consistent superiority in minimizing large prediction errors. The \gls{mae} plot (middle panel) provides a more balanced view of average prediction accuracy, with values ranging from 0 to 0.4. The linear nature of \gls{mae} shows that while absolute differences between models are smaller than in MSE, the relative performance ranking remains consistent. This suggests that our model's advantages are not merely due to outlier handling but represent genuine improvements in average prediction accuracy. The \gls{rmse} plot (bottom panel) combines the sensitivity of MSE to large errors with the interpretability of \gls{mae}'s scale. Values range from 0 to 0.5, and the patterns closely mirror those in the MSE plot but with a compressed scale. This metric confirms that our model's superior performance is robust across different error measurement approaches.

Our Model (Red) consistently maintains the smallest error region across all metrics, with particularly strong performance on challenging participants (0, 8, 9, 10, 11, 12, 13, 14, 15). The compact radar area indicates robust generalization across diverse sleep patterns.

These results validate our model's effectiveness in handling the inherent variability in sleep patterns across individuals while maintaining superior prediction accuracy compared to state-of-the-art baselines.

 \begin{figure*}
    \centering
    \includegraphics[width=0.8\linewidth]{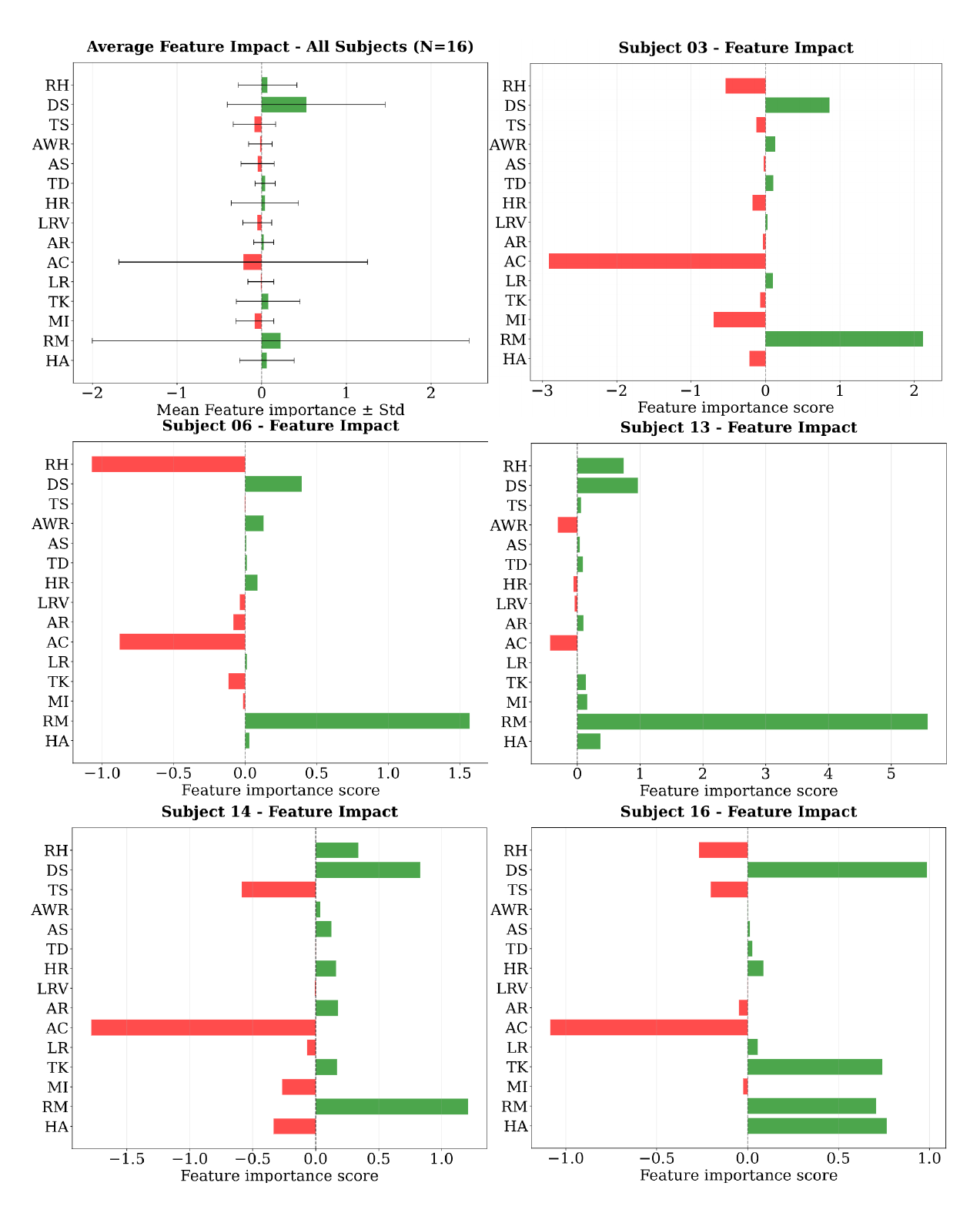}
    \caption{Features importance analysis with impact direction by \gls{shap} analysis}
    \label{fig:explain}
\end{figure*}

\subsection{Explainable model for sleep forecast}

The results reveal distinct patterns of feature importance across participants, demonstrating both individual variability and common influential factors in sleep quality prediction, as shown in Figure~\ref{fig:explain}.

\textbf{Average Feature Importance Analysis}: 
The aggregated analysis of all 16 subjects reveals that \textbf{deep sleep seconds (DS)} emerges as the strongest positive predictor of sleep quality. This finding validates the fundamental importance of deep sleep stages in determining overall sleep quality and restorative function. The robust positive impact of DS across the entire subject population underscores the critical role of slow-wave sleep in sleep quality assessment. Additionally, \textbf{resting heart rate (RH)} demonstrates a consistent moderate positive influence (mean $\approx$ +0.4), reinforcing the relationship between cardiovascular fitness and sleep quality outcomes.

\subsubsection{Clinical Implications of Feature Importance}

The analysis provides several clinically relevant insights:

\begin{enumerate}
\item \textbf{Restlessness as a Primary Indicator}:The moderate positive influence of sleep stage durations (\texttt{Deep Sleep Seconds}, \texttt{Light Sleep Seconds}, \texttt{Rem Sleep Seconds}) confirms that our model appropriately values sleep architecture. The balanced contributions suggest the model considers overall sleep composition rather than overemphasizing any single stage.

\item \textbf{Individual Variability}: 
The individual subject analyses reveal substantial heterogeneity in feature response patterns, with response scores ranging from Subject 13's exceptional \textbf{restless moment count (RM)} impact of $+5.0$ to Subject 03's severe \textbf{awake count (AC)} sensitivity of $-2.5$. Subject 03 exhibits the most extreme sensitivity to sleep fragmentation, with AC demonstrating an exceptionally strong negative impact ($-2.5$) while RM shows a strong positive impact ($+1.8$), and \textbf{resting heart rate (RH)} unexpectedly shows a negative impact ($-0.8$), contrasting with the general population pattern. Subject 06 demonstrates a unique cardiovascular-sleep interaction pattern, with RH showing the strongest negative impact ($-1.0$) among all measured features, while RM exhibits a high positive impact ($+1.4$), suggesting interventions targeting both cardiovascular health and sleep environment optimization would be beneficial. Subject 13 presents the most remarkable profile, with RM demonstrating an unprecedented positive impact approaching $+5.0$, indicating that natural sleep movements are critically important for this individual's sleep quality maintenance. Subject 14 exhibits a balanced response pattern with AC showing substantial negative impact ($-1.4$), while RM and \textbf{deep sleep seconds (DS)} both demonstrate strong positive influences ($+1.0$ each), and \textbf{total steps (TS)} shows notable negative impact ($-0.7$). Subject 16 demonstrates a distinctive activity-responsive pattern where \textbf{total kilocalories (TK)}, \textbf{highly active seconds (HA)}, and RM all exhibit substantial positive impacts ($+0.8$, $+0.8$, and $+0.7$ respectively), while AC shows strong negative influence ($-1.0$) and both RH and TS show moderate negative impacts.

The differences in feature response patterns across subjects underscore the critical importance of personalized sleep medicine approaches, as this 7.5-point range in individual feature responses far exceeds population-level variations observed in the aggregated analysis. The identification of distinct phenotypes—including sleep fragmentation sensitive (Subject 03), cardiovascular-arousal sensitive (Subject 06), movement-dependent (Subject 13), multi-factor balanced (Subject 14), and activity-responsive (Subject 16) profiles—provides a framework for developing targeted interventions. 

\end{enumerate}

\subsubsection{Model Behavior Validation}

The interpretability results demonstrate several positive aspects of our model's behavior:

\begin{enumerate}
\item \textbf{Physiological Coherence}: The model assigns importance to physiologically meaningful features. Respiratory values during sleep and waking states show consistent but moderate importance, indicating the model captures subtle physiological patterns without overfitting to noise.

\item \textbf{Absence of Spurious Correlations}: Features with low clinical relevance are appropriately de-emphasized, suggesting the model has learned meaningful rather than spurious relationships.

\item \textbf{Bidirectional Effects}: The presence of both positive and negative values for different features indicates the model captures complex, relationships. For instance, while increased deep sleep positively influences predictions, excessive restlessness negatively impacts them, reflecting real-world sleep dynamics.
\end{enumerate}

\subsubsection{Practical Applications}

The interpretability framework enables several practical applications:

\begin{enumerate}
\item \textbf{Personalized Sleep Interventions}: By identifying individual-specific important features, clinicians can tailor interventions. The high ranking of daily activity in participant 03, for example, could indicate that an intervention targeting activity might be more beneficial for this individual.

\item \textbf{Model Trust and Adoption}: The ability to explain predictions using clinically meaningful features enhances trust in the model's recommendations, facilitating adoption in clinical settings.

\item \textbf{Feature Engineering Guidance}: The importance rankings can guide future feature engineering efforts, potentially focusing on more detailed restlessness metrics or sleep fragmentation indices.
\end{enumerate}

This interpretability analysis transforms our deep learning model from a ``black box'' into a transparent tool that provides actionable insights for both clinicians and patients.

\section{Conclusions and Limitations}


This study presents a comprehensive deep learning framework for sleep quality forecasting that addresses key challenges in explainable and personalized sleep management. Our two-stage domain adaptation model consistently outperforms state-of-the-art baselines across all evaluation metrics, achieving trend correlations exceeding 0.8 for most participants and trend direction accuracies above 75\%. These results demonstrate the model's robust capability in capturing both short-term fluctuations and long-term sleep quality patterns.

Our model could inform users and clinicians about future potential deterioration, thereby facilitating individualized and timely interventions. The model's ability to maintain performance across extended periods (up to 170 days) and diverse sleep patterns validates its potential for real-world deployment in continuous sleep monitoring systems. Through analysis, we identified Deep Sleep Seconds as the most influential predictor across participants, while revealing significant individual variations in feature importance. This interpretability framework transforms our model from a black box into a transparent tool that provides actionable insights for personalized sleep interventions.

Our model demonstrates remarkable consistency across 16 participants with diverse sleep patterns, from highly regular to extremely variable. The smooth transition between validation and test sets, coupled with adaptive uncertainty quantification, confirms the model's generalization capability and clinical reliability. Furthermore, our findings emphasize that effective sleep quality prediction requires individualized approaches, as substantial inter-participant variations in both model performance and feature importance highlight the inadequacy of one-size-fits-all solutions in sleep management.

Despite promising results, several limitations merit consideration. First, the model relies solely on wearable sensor data without incorporating environmental factors (temperature, noise), lifestyle variables (caffeine, work schedules), or psychological states that could enhance prediction accuracy for irregular patterns. Second, while demonstrating strong technical performance, prospective clinical validation is needed to assess real-world impact on sleep interventions. The study is limited by using Garmin-derived sleep scores that may not represent a clinically recognized sleep quality measure, though the approach is applicable to other sleep assessment methods. Future work should focus on adaptive models for sudden pattern changes, continuous learning frameworks, multi-modal data fusion, and standardized benchmarks to advance toward clinically impactful solutions.


\bibliographystyle{unsrt}  
\bibliography{ref}         

\end{document}